\documentclass[10pt,twocolumn,letterpaper]{article}

\usepackage{cvpr}              %

\usepackage{multirow}
\usepackage{tabularray}
\usepackage{gensymb}

\definecolor{cvprblue}{rgb}{0.21,0.49,0.74}
\usepackage[pagebackref,breaklinks,colorlinks,allcolors=cvprblue]{hyperref}

\usepackage[accsupp]{axessibility}

\title{Glossy Object Reconstruction with Cost-effective Polarized Acquisition}

\author{Bojian Wu$^1$  \quad  Yifan Peng$^{2,*}$  \quad  Ruizhen Hu$^3$  \quad  Xiaowei Zhou$^{1,*}$ 
\\ $^1$Zhejiang University  \quad  $^2$The University of Hong Kong  \quad $^3$Shenzhen University}

\begin{document}

\twocolumn[{%
\renewcommand\twocolumn[1][]{#1}%
\maketitle
\centering
\includegraphics[width=\textwidth]{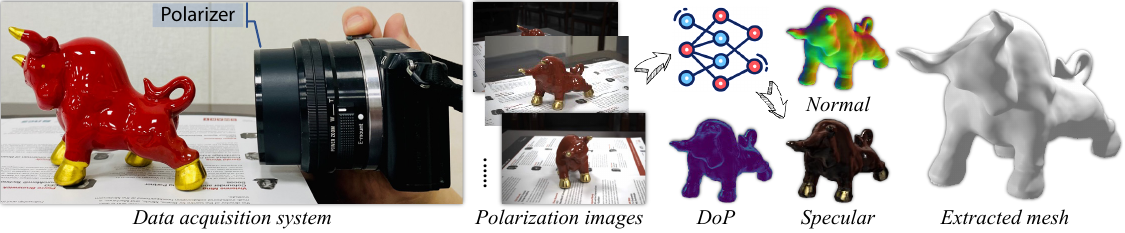}
\vspace{-13pt}
\captionof{figure}{We build a cost-effective data acquisition system for capturing multi-view polarization images, where a linear polarizer is mounted in front of the off-the-shelf RGB camera and a single image per-view with unknown angle of the polarizer is captured, which eliminates the need for precise alignment. For objects with a hybrid of ceramics (tummy) and metal (feet), we can still nicely recover the specular components and estimate the polarimetric states, directly leading to high-fidelity geometry. \vspace{3em}}
\label{fig:teaser}
}]

\renewcommand{\thefootnote}{$*$}\footnotetext{Corresponding authors.}
\renewcommand{\thefootnote}{\arabic{footnote}}

\begin{abstract}
The challenge of image-based 3D reconstruction for glossy objects lies in separating diffuse and specular components on glossy surfaces from captured images, a task complicated by the ambiguity in discerning lighting conditions and material properties using RGB data alone. While state-of-the-art methods rely on tailored and/or high-end equipment for data acquisition,  which can be cumbersome and time-consuming, this work introduces a scalable polarization-aided approach that employs cost-effective acquisition tools. By attaching a linear polarizer to readily available RGB cameras, multi-view polarization images can be captured without the need for advance calibration or precise measurements of the polarizer angle, substantially reducing system construction costs. The proposed approach represents polarimetric BRDF, Stokes vectors, and polarization states of object surfaces as neural implicit fields. These fields, combined with the polarizer angle, are retrieved by optimizing the rendering loss of input polarized images. By leveraging fundamental physical principles for the implicit representation of polarization rendering, our method demonstrates superiority over existing techniques through experiments in public datasets and real captured images on both reconstruction and novel view synthesis.
\end{abstract}

\section{Introduction}\label{sec:intro}

3D reconstruction has been a long-standing topic in the graphics and vision communities. 
State-of-the-art methods are mostly designed for opaque surfaces with the Lambertian reflectance model and may perform sub-optimally in non-Lambertian scenes~\cite{FRT18,DRT20}, posing a challenge for both acquisition systems and reconstruction algorithms.

In particular, to deal with glossy or specular regions, except for painting with diffuse coats, specially-tailored devices are often required for recording the controlled environmental illumination and/or reflective lighting conditions. An alternative approach explores polarization cues, referred to as \textit{Shape-from-Polarization (SfP)}~\cite{cui2017polarimetric,zhao2022polarimetric,fukao2021polarimetric}, as polarization properties are closely related to surface normals. Moreover, diffuse and specular reflectances exhibit different polarimetric statuses, with the specular being more polarized than the diffuse and their polarization angles being orthogonal. These physical insights can be valuable for algorithms.

The existing optimization-based \textit{SfP} methods face challenges when processing irregular triangles or non-manifold mesh, that could be largely overcome by incorporating neural implicit surfaces. Dave~\etal~\cite{dave2022pandora} propose the first implementation that integrates polarization cues into neural radiance fields. It should be noted, however, that this approach requires an expensive polarization camera for data acquisition to obtain full polarization states, such as Stokes vectors, as supervision for network training. 
In contrast, we argue that, an off-the-shelf RGB camera equipped with a linear polarizer can already effectively acquire the required data, thereby greatly reducing the system cost.

Our approach employs a single captured polarization image per view as input and builds upon the polarimetric BRDF (pBRDF) model~\cite{baek2018simultaneous}, which explicitly 
models the relation between polarization states of outgoing radiance and surface properties.
To represent the object's geometry, we utilize the neural implicit surface, that enables us to query the signed distance values and surface normals at any scene points. 
With scene coordinates, surface normals, and view directions as input, we employ separate radiance networks to represent the diffuse and specular radiances. These radiances form the basis for computing polarization states, which are depicted by the Stokes vectors and computed using the pBRDF model.
Finally, the polarized images are rendered using volume rendering given the Stokes vectors at sampled scene points and the angle of polarizer. 
By minimizing the rendering loss between the rendered polarized images and the input polarized images, we recover neural radiance fields and surface properties.
Importantly, the polarizer angle, which is typically unknown without complex calibration procedures, can be optimized along with the networks.
Results tested on both public datasets and real captured data (Sec.~\ref{sec:exp}) demonstrate the effectiveness and robustness of our approach (see the example in Fig.~\ref{fig:teaser}). The main contributions are as follows: 
\vspace{3pt}
\begin{itemize}
    \item We devise an cost-effective setup for acquiring polarization images by integrating an off-the-shelf RGB camera with a linear polarizer, eliminating the need for labor-intensive calibration and reducing the overall cost.
    \item We are the first to leverage a single polarization image per view, in conjunction with neural radiance fields and fundamental physical principles, to enable the end-to-end polarization rendering.
    \item Experimental results demonstrate that our method well handles non-Lambertian components, leading to high fidelity geometry and radiance decomposition.
\end{itemize}

\section{Related Work}\label{sec:rw}

We will next discuss only the methods of radiance decomposition and geometry recovery for glossy/specular objects using Neural Radiance Fields (NeRF)~\cite{mildenhall2020nerf}.

\begin{figure*}[t!]
    \centering
    \includegraphics[width=\textwidth]{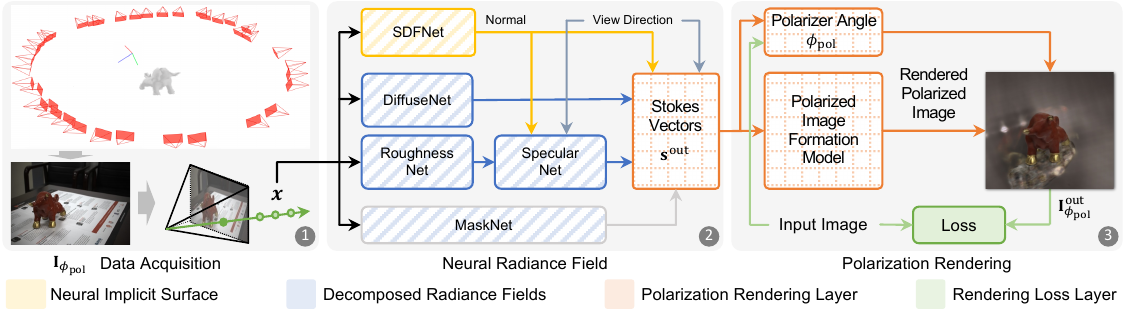}
    \caption{\textbf{Overview of neural glossy object reconstruction with polarization cues.} Our method consists of three main steps (1--3): data acquisition, neural radiance field-based representation, and polarization rendering.
    This work employs neural rendering techniques in conjunction with the fundamental principles of polarization to generate a polarized image. 
    These coupled modules allow for acquiring only one single polarization image at each viewing angle and then recover geometry and material properties through the optimization of rendering loss.
    Components marked with upward diagonal strips, such as $\mathbf{DiffuseNet}$ and $\mathbf{SpecularNet}$, are optimized during training, while those with grid checker patterns are calculated using corresponding equations.}
    \label{fig:network}
\end{figure*}

\paragraph{Glossy and specular surface reconstruction.} Recent attempts such as Zhang~\etal~\cite{zhang2021physg} and Boss~\etal~\cite{boss2021neural} aim to address this ill-posed problem by decomposing the specular reflectance with the estimated BRDF. Guo~\etal~\cite{guo2022nerfren} split a scene into transmitted and reflected components, that are modeled with separate neural radiance fields. Verbin~\etal~\cite{verbin2022ref} consider spatially-varying scene properties and parameterize the outgoing radiance with the directional encoding of the reflected radiance. Yan~\etal~\cite{yan2023nerf} extend this idea to dynamic scenes with a masked guided deformation field. Xu~\etal~\cite{xu2021scalable} leverage an image-based rendering pipeline to reconstruct depth and reflection, and then select adjacent views for plausible coherent renderings. Kopanas~\etal~\cite{kopanas2022neural} propose a neural warp field to model catacaustic trajectories of reflections, which enables efficient point splatting-based rendering for complex specular effects.
Although better rendering effects can be obtained, these methods often ignore the quality of geometry~\cite{yang2022ps,zhu2023vdn}. Reconstruction results can be refined by balancing the importance of regions with different surface properties, such as adaptive reflection-aware photometric loss~\cite{ge2023ref}. Liu~\etal~\cite{liu2023nero} propose to utilize two individual networks to encode the radiance of direct and indirect lights, respectively, which are selected subject to an estimated occlusion probability during rendering. Such a representation efficiently accommodates accurate surface reconstruction of reflective objects.

\paragraph{Shape from Polarization (SfP).} Traditional SfP requires consideration of multi-view consistency, and constraints on the continuity and smoothness of the mesh surface to address the singularities in angle and phase caused by polarization, for better reconstruction~\cite{cui2017polarimetric,fukao2021polarimetric,zhao2022polarimetric,shao2024polarimetric,zhao2024polarimetric}.
Recent years have witnessed significant advancements of volume rendering based methods in resolving the shape~\cite{li2024neisf,cao2025nerf,han2024nersp,chen2024pisr,yang2024gnerp,tiwari2024ss,peters2023pcon,kim2023nespof}. To be specific, Dave~\etal~\cite{dave2022pandora} propose the pioneering work and first incorporate polarization cues into the neural radiance field and train the network using polarization states instead of original color information. This approach naturally facilitates decomposition of radiance into diffuse and specular components, leading to improved geometries.
However, accurately characterizing  polarization information often requires precise rotation and calibration of the polarizer mounted in front of the camera, which can be a tedious task and limits practical utilization. Although emerging snapshot polarization image sensors (e.g., Sony IMX250MZR on-chip polarizer~\cite{sony}), allow for the acquisition of multi-directional polarized images in a single capture, the cost of such devices makes them impractical for personal use. To bypass the drawbacks of both approaches, we utilize only an RGB camera and a linear polarizer to establish an efficient yet low-cost acquisition scheme, eliminating the need for tedious pre-calibration.

\section{Method}\label{sec:method}

\subsection{Overview of Reconstruction Pipeline}
We aim to reconstruct the geometry and appearance of a glossy object from a set of posed polarization images $\{\mathbf{I}^k_{\phi_\text{pol}}\}$, where the angle of the polarizer filter $\phi_\text{pol}$ is unknown. The entire pipeline, depicted in Fig.~\ref{fig:network}, consists of three main steps. To commence, we randomly select multiple camera poses surrounding the target object and capture a single polarization image $\mathbf{I}_{\phi_\text{pol}}$ at each view with our low-cost data acquisition system, as shown in Fig.~\ref{fig:teaser}. Next, in alignment with prior study~\cite{dave2022pandora}, we employ VolSDF~\cite{yariv2021volume} and Ref-NeRF~\cite{verbin2022ref} as the fundamental blocks for modeling the neural implicit surface and decomposed radiances. Then, we harness the polarimetric BRDF model to accurately estimate Stokes vectors $\mathbf{s}^\text{out}$. Furthermore, we introduce an end-to-end polarization rendering layer, which first estimates the polarizer's angle $\phi_\text{pol}$ and then incorporates physical rules to render a polarized image $\mathbf{I}^\text{out}_{\phi_\text{pol}}$, which is compared with the captured ground-truth for loss calculation.

As in Fig.~\ref{fig:network}, our method utilizes a polarization image $\mathbf{I}_{\phi_\text{pol}}$ as the input and initiates by sampling a collection of 3D locations along each camera ray. These locations are processed through a coordinate-based neural implicit surface module, facilitating the estimation of signed distances and surface normals. Along with view directions, separate radiance networks are employed to determine the diffuse and specular components. This separation allows us to effectively handle the non-Lambertian properties exhibited by the surface. Combined with the polarimetric BRDF model, the outgoing Stokes vectors $\textbf{s}^\text{out}$ can be obtained, which lay the foundation for polarization-based rendering. The details on these methods can be found in supplementary materials.

Next, we present a differentiable processing pipeline to estimate the polarizer's angle $\phi_\text{pol}$, eliminating the need for precise polarization angle measurements and facilitating the implicit rendering of desired polarized images $\mathbf{I}^\text{out}_{\phi_\text{pol}}$ for loss calculation. Subsequently, we provide a comprehensive analysis of the fundamental principles of polarization and its application in aiding the reconstruction and radiance decomposition in Sec.~\ref{subsec:pol_rendering}. Moreover, we illustrate the rationale behind the efficacy of using a single polarization image per view to achieve our goals and elucidate the distinctions between this approach and prior methodologies in Sec.~\ref{subsec:theory}.

\subsection{Polarization-empowered Rendering}\label{subsec:pol_rendering}

In this approach, we take the estimated outgoing Stokes vector $\mathbf{s}^\text{out}$ as input, which characterizes the polarization state of light and is represented by a four-dimensional vector $[s_0,s_1,s_2,s_3]$. From this, we calculate the fundamental polarization information as follows:
\begin{equation}\label{eq:pol_info}
    \mathbf{I}_\text{un} = \frac{1}{2}s_0,
    ~\rho = \frac{\sqrt{s_1^2+s_2^2}}{s_0},
    ~\phi = \frac{1}{2}\text{arctan2}(s_2,s_1),
\end{equation}
where $\rho$ is the degree of polarization (DoP), $\phi$ is the angle of polarization (AoP), and $\mathbf{I}_\text{un}$ is the unpolarized intensity.

On the one hand, the polarized intensity $\mathbf{I}_{\phi_\text{pol}}$ (i.e., the captured image) exhibits sinusoidal variation with the rotation angle of the polarizer $\phi_\text{pol}$, as shown below:
\begin{equation}
\label{eq:pol}
\mathbf{I}_{\phi_\text{pol}} = \mathbf{I}_\text{un}\left( 1 + \rho \cos (2\phi - 2\phi_\text{pol}) \right).
\end{equation}
Using Eq.~\ref{eq:pol_info}, the only unknown variable $\phi_\text{pol}$ can be easily solved given $\mathbf{s}^\text{out}$ and $\mathbf{I}_{\phi_\text{pol}}$.

Moreover, Mueller matrices are only valid in the aligned reference coordinate system when considering the light passing through a polarizer. Therefore, for a linear polarizer with a rotation angle of $\phi_\text{pol}$, its Mueller matrix must be deduced according to~\cite{collett1992polarized}:
\begin{equation}
    \label{eq:mueller_lp}
     \mathbf{M_{\phi_\text{pol}}} = \mathbf{R_{\phi_\text{pol}}^T} \mathbf{M_{LP}} \mathbf{R_{\phi_\text{pol}}},
\end{equation}
where $\mathbf{R}_{\phi_\text{pol}}$ is the rotation matrix and $\mathbf{M_{LP}}$ is the Mueller matrix of an ideal linear polarizer with the horizontal transmission. Both are defined as follows:
\begin{equation}
\resizebox{0.92\hsize}{!}{$%
  \mathbf{R_{\phi_\text{pol}}} = \begin{bmatrix}
    1 & 0 & 0 & 0 \\ 
    0 &  \cos(2\phi_\text{pol}) & \sin(2\phi_\text{pol}) & 0 \\ 
    0 & -\sin(2\phi_\text{pol}) & \cos(2\phi_\text{pol}) & 0 \\ 
    0 & 0 & 0 & 1
    \end{bmatrix},~
 \mathbf{M_{LP}} = \begin{bmatrix}
     0.5 & 0.5 & 0 & 0 \\ 
     0.5 & 0.5 & 0 & 0 \\ 
     0 & 0 & 0 & 0 \\ 
     0 & 0 & 0 & 0
    \end{bmatrix}.
$}%
\end{equation}
Accordingly, passing/modulating through a linear polarizer, the outgoing Stokes vector $\mathbf{s}^\text{out}$ can be transformed by:
\begin{equation}
    \label{eq:lp_phi_pol}
    \mathbf{s}^\text{out}_{\phi_\text{pol}} = \mathbf{M_{\phi_\text{pol}}} \mathbf{s}^\text{out} = \mathbf{R_{\phi_\text{pol}}^T} \mathbf{M_{LP}} \mathbf{R_{\phi_\text{pol}}} \mathbf{s}^\text{out}.
\end{equation}
Then, the final rendered polarized image is denoted by:
\begin{equation}
\mathbf{I}^\text{out}_{\phi_{\text{pol}}} = \frac{1}{2} \mathbf{s}^\text{out}_{\phi_\text{pol}}[0],
\end{equation}
where $\mathbf{s}^\text{out}_{\phi_\text{pol}}[0]$ is the first element of Stokes vector.

\paragraph{Loss function.}
In order to describe the polarization status in the region of interest (RoI) and reduce the background noise, we apply a coordinate-based network to predict the soft mask $m(\mathbf{x})$ of each sampled point $\mathbf{x}$ on the camera ray. 
Therefore, the complete loss function consists of three components with balancing weights denoted as follows:
\begin{equation}
    \mathcal{L} = \mathcal{L}_\text{rgb}  + \mathcal{L}_\text{mask} + 0.1 \mathcal{L}_\text{eikonal}.
    \label{eq:loss_fn}
\end{equation}
The RGB loss $\mathcal{L}_\text{rgb}$ describes the discrepancies between the rendered polarized image $\mathbf{I}_{\phi_\text{pol}}^\text{out}$ and the captured image $\mathbf{I}_{\phi_\text{pol}}$ using $\ell_1$ loss. The loss is masked with the ground-truth mask to reduce the noise from surrounding environment. The predicted mask is supervised by the ground-truth mask with the binary cross entropy loss $\mathcal{L}_\text{mask}$. In addition, we introduce the eikonal loss $\mathcal{L}_\text{eikonal}$~\cite{gropp2020implicit} to regularize the network to learn a valid signed distance field (SDF).

\subsection{Theoretical Analysis}\label{subsec:theory}
Our method aims to retrieve not only geometric and polarization information but also the  polarizer's angle from multi-view images, requiring only one polarization image per view,
which presents us with more unknown variables to address within a reduced set of limitations.

As aforementioned, we utilize the polarimetric BRDF model to express the Stokes vector $\mathbf{s}^\text{out}$ as a linear combination of polarized diffuse and specular counterparts, respectively. Here, we focus solely on the radiance component:
\begin{equation}
\label{eq:rad}
\mathbf{I}^\text{out} = (\mathbf{n}\cdot\mathbf{i}) \left( f_d(\mathbf{i}, \mathbf{n}, \mathbf{v}) + f_s(\mathbf{i}, \mathbf{n}, \mathbf{v},\eta)\right) L_i,
\end{equation}
where $\mathbf{i}$, $\mathbf{n}$, $\mathbf{v}$ and $\eta$ denote the incident lighting direction, normal, viewing direction, and roughness. $L_i$ is incident illumination and is usually defined as white light ($L_i=1.0$).

The diffuse reflectance $f_d$ pertains to light that enters the subsurface, scatters, and subsequently transmits back in the direction of observation. The specular reflectance $f_s$ models both specular lobe and spike, which are defined below:
\begin{equation}
\begin{split}
    f_d &= k_d T(\mathbf{v}, \mathbf{n})T(\mathbf{i}, \mathbf{n}), \\ f_s & = k_s W(\mathbf{i}, \mathbf{n}, \mathbf{v}, \eta) R(\mathbf{h}, \mathbf{v}),
\end{split}
\end{equation}
where $W=\frac{DG}{4(\mathbf{n}\cdot\mathbf{o})}$, and all other parameters are defined in the same manner as outlined in~\cite{baek2018simultaneous}.

The Fresnel coefficients $T$ and $R$ at polarization filter angle $\phi_\text{pol}$ are represented by:
\begin{equation}
    T_{\phi_\text{pol}} = \frac{T_p + T_s}{2} \:+\: \rho_t\frac{T_p-T_s}{2}\cos(2\phi_t - 2\phi_\text{pol}),
\end{equation}
\begin{equation}
    R_{\phi_\text{pol}} = \frac{R_s + R_p}{2} \:+\: \rho_r\frac{R_s-R_p}{2}\cos(2\phi_r - 2\phi_\text{pol}),
\end{equation}
where the subscriptions $p$ and $s$ indicate the components parallel and perpendicular to the reflection plane, while $\rho_t$ and $\rho_r$ represent the degree of linear polarization for transmittance and reflection respectively, $\phi_t$ and $\phi_r$ correspond to the angle of polarization of transmission and reflection.

Ultimately, the output estimated radiance $\mathbf{I}^\text{out}$ (Eq.~\ref{eq:rad}) at the polarization filter angle $\phi_\text{pol}$ can be expressed as follows:
\begin{equation}
\label{eq:i_est}
\begin{split}
\mathbf{I}^\text{out}_{\phi_\text{pol}} = (\mathbf{n}\cdot\mathbf{i}) & \left( k_d T(\mathbf{v}, \mathbf{n},\phi_\text{pol})T(\mathbf{i}, \mathbf{n}) \: + \right. \\ 
&~ \left. k_s W(\mathbf{i}, \mathbf{n}, \mathbf{v}, \eta) R(\mathbf{h}, \mathbf{v}, \phi_\text{pol}) \right) L_i.
\end{split}
\end{equation}
In our implementation, the incident direction $\mathbf{i}$ of the light is approximated as the reflected direction of $\mathbf{v}$, thereby aligning the half vector $\mathbf{h}$ with the normal direction. Consequently, the unknown variables in Eq.~\ref{eq:i_est} are limited to $\mathbf{n}$ (2 unknowns, parameterized in spherical coordinates), $k_d$ (3 unknowns), $k_s$ (3 unknowns), $\eta$ (1 unknown), and $\phi_\text{pol}$ (1 unknown), totaling 10 unknowns. It is worth noting that, except for $\phi_\text{pol}$, the remaining variables represent intrinsic material properties of the object and are fully disentangled within this material model. These variables remain consistent for the same spatial point,  irrespective of the viewing angle. The view dependency of color provides 3 separate constraints (R, G and B) for each view, implying that only \emph{four} views are sufficient to render the problem over-determined, eventually forming 12 independent equations.

\paragraph{Distinction to prior works.} In contrast to the well-established polarization method, i.e., PANDORA~\cite{dave2022pandora}, our method necessitates the acquisition of one single polarization image at each viewing angle. We employ the proposed end-to-end rendering framework and enhance geometric and material reconstruction through optimization of the rendering loss function.
Comparing with conventional non-polarization solutions, such as VolSDF~\cite{yariv2021volume}, our method stands out in rendering out the higher-quality surface reconstruction.
While multi-view consistency assumptions tend to break down when dealing with glossy surfaces in certain scenes, 
our polarization setup allows for the effective modeling of RGB information from various perspectives through polarization rendering, as denoted by $\mathbf{I}_{\phi_\text{pol}}^\text{out}$ earlier. 
This unique representation seamlessly integrates both the object's normal vector and material properties, facilitating the deduction of geometric characteristics and material properties within a unified framework. By progressively enhancing the accuracy of $\mathbf{I}_{\phi_\text{pol}}^\text{out}$ through the minimization of rendering loss, we implicitly refine the accuracy of normal vector and subsequently elevate the quality of geometry.

\begin{figure}[t!]
\centering
\includegraphics[width=\columnwidth]{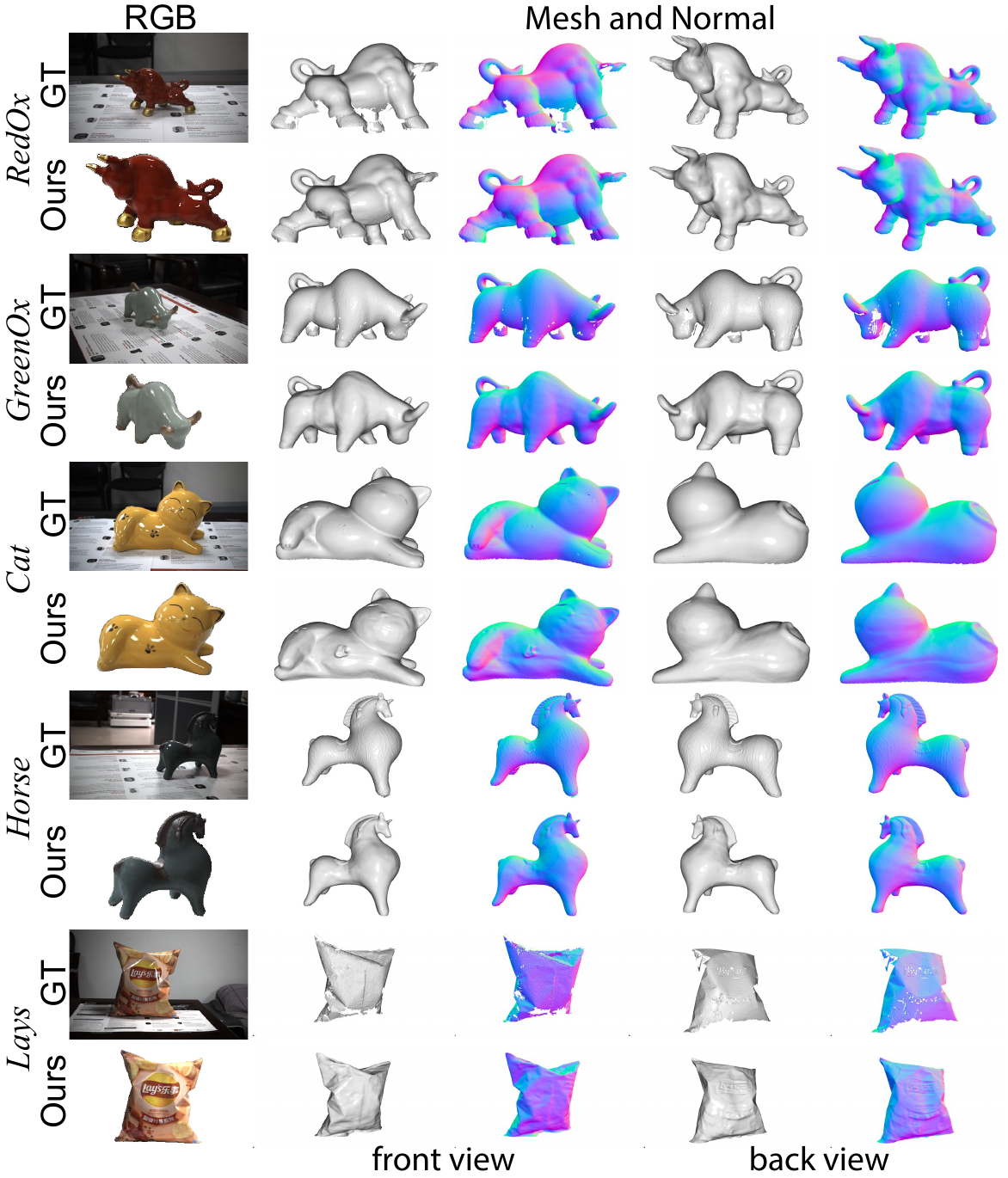}
\caption{\textbf{Qualitative results of captured datasets.} For each scenario, the top row shows the input reference image, ground-truth mesh (obtained by painting and scanning), and corresponding normals; the bottom row demonstrates our resolved results, including the rendered image and extracted mesh.}
\label{fig:results-red_ox}
\end{figure}

\section{Experiment}\label{sec:exp}

\subsection{Datasets and Results}\label{subsec:datasets}
To meet our requirements, we build a simple data acquisition system using off-the-shelf products, which includes an RGB camera (SONY A6400 with 4K resolution) and a linear polarizer, as shown in Fig.~\ref{fig:teaser}. We select several complex objects with varying materials, such as ceramics, metal, and plastic, see examples in Figs.~\ref{fig:teaser} (\textit{RedOx}) and ~\ref{fig:results-red_ox} (\textit{GreenOx}, \textit{Cat}, \textit{Horse} and \textit{Lays}). In practice, we fix the orientation of polarizer across all the captured views and hold the device to collect images approximately evenly around the object, see example camera poses in Fig.~\ref{fig:network}. The multi-view images are captured under uncontrolled indoor lighting environments, and about 40 images are enough for each object. In all cases, we first downsample the image by a factor of 4 and apply COLMAP~\cite{schoenberger2016sfm} to obtain the initial poses.

Results tested on \textit{RedOx} model and others are shown in Figs.~\ref{fig:teaser} and ~\ref{fig:results-red_ox}. Note that, for a variety of different materials (ceramics, metal, etc.), with varying lighting conditions, our method still recovers the surface geometry reasonably well.
Moreover, the fact that polarization cues behave differently for the diffuse and specular components greatly aid in understanding material properties and facilitating radiance decomposition, which is an inherently ill-posed problem. As depicted in presented examples, our results reasonably separate the diffuse and specular components. Additionally, the estimated polarimetric cues align with our intuition, i.e., the AoP is orthogonal for the diffuse and specular components, while the DoP is higher for the specular regions.

\begin{figure}[tp!]
\centering
\includegraphics[width=\columnwidth]{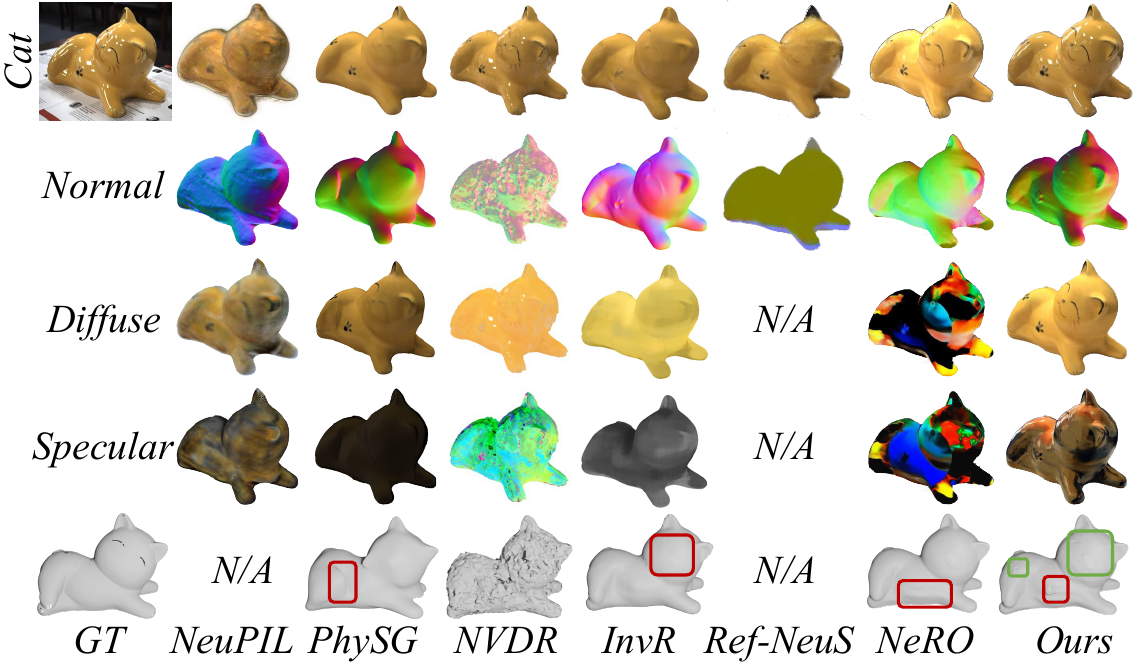}
\caption{\textbf{Qualitative comparison with SOTA methods.} Our approach excels in reconstructing intricate features such as beard and tail segments, due to the advantage of the polarization information.}
\label{fig:comp-sota}
\end{figure}

\begin{table*}[t]
\caption{\textbf{Quantitative assessment of rendering and reconstruction quality.}
To ensure a fair comparison in the 3D reconstruction quality, all models are normalized to the unit sphere. Note that, we do not directly compare with NeuralPIL and Ref-NeuS, as they fail to produce valid geometry in several cases, as evident in Fig.~\ref{fig:comp-sota}. Nevertheless, with the incorporation of polarization cues, our method consistently achieves the best results.}
\label{tab:quality}
\vspace{-6pt}
\centering
\resizebox{\textwidth}{!}{
\begin{tabular}{c|ccc|ccc|ccc|ccc|ccc}
\hline
\multirow{2}{*}{} & \multicolumn{3}{c|}{\textit{RedOx}} & \multicolumn{3}{c|}{\textit{GreenOx}} & \multicolumn{3}{c|}{\textit{Cat}} & \multicolumn{3}{c|}{\textit{Horse}} & \multicolumn{3}{c}{\textit{Lays}} \\ \cline{2-16} 
 & PSNR~$\uparrow$ & SSIM~$\uparrow$ & CD~$\downarrow$ & PSNR & SSIM & CD & PSNR & SSIM & CD & PSNR & SSIM & CD & PSNR & SSIM & CD \\ \hline
NeuralPIL & 22.47 & 0.9378 & - & 23.51 & 0.9301 & - & 22.36 & 0.8626 & - & 21.27 & 0.9049 & - & 20.93 & 0.8657 & - \\
PhySG & 16.42 & 0.9737 & 2.36e-2 & 18.39 & 0.9822 & 1.43e-2 & 16.32 & 0.9513 & 1.48e-3 & 16.59 & 0.9518 & 1.31e-3 & 17.41 & 0.9569 & 2.66e-3 \\
NVDiffRec & 30.86 & 0.9639 & 0.3005 & 30.66 & 0.9862 & 0.2638 & 23.61 & 0.9614 & 0.5936 & 27.15 & 0.9590 & 0.1315 & 29.31 & 0.9693 & 0.1152 \\
InvRender & 22.47 & 0.9631 & 2.28e-2 & 27.32 & 0.9758 & 1.78e-2 & 22.32 & 0.9510 & 1.82e-3 & 24.92 & 0.9464 & 1.13e-3 & 25.61 & 0.9673 & 1.21e-3 \\
Ref-NeuS & 27.21 & 0.8562 & - & 27.35 & 0.8528 & - & 23.27 & 0.8464 & - & 23.45 & 0.8562 & - & 27.28 & 0.91753 & - \\
NeRO & 19.88 & 0.8503 & 2.04e-3 & 16.98 & 0.5972 & 1.08e-3 & 24.51 & 0.8039 & 9.31e-3 & 22.22 & 0.8294 & 1.20e-3 & 26.68 & 0.9256 & 1.04e-3 \\
Ours (diffuse) & 25.03 & 0.9683 & 1.06e-3 & 28.24 & 0.9860 & 7.99e-4 & 24.39 & 0.9465 & 5.91e-3 & 22.43 & 0.8996 & 7.88e-4 & 28.52 & 0.9457 & 3.50e-3 \\
Ours (w/o pol) & 26.29 & 0.9662 & 3.01e-3 & 30.77 & 0.9738 & 7.14e-4 & 23.84 & 0.9343 & 1.39e-3 & 25.84 & 0.9566 & 6.76e-4 & 24.04 & 0.9520 & 2.76e-3 \\
Ours & \textbf{30.88} & \textbf{0.9774} & \textbf{2.23e-4} & \textbf{31.02} & \textbf{0.9883} & \textbf{1.17e-4} & \textbf{24.83} & \textbf{0.9696} & \textbf{9.88e-5} & \textbf{27.97} & \textbf{0.9606} & \textbf{2.07e-4} & \textbf{30.82} & \textbf{0.9780} & \textbf{1.01e-3} \\ \hline
\end{tabular}
}
\end{table*}

\subsection{Assessments against Counterparts}

\paragraph{Comparisons with non-polarization methods.}
We have conducted a comparison of our approach with several state-of-the-art radiance decomposition and surface reconstruction methods. For instance, as depicted in Fig.~\ref{fig:comp-sota}, NeuralPIL~\cite{boss2021neural} and PhySG~\cite{zhang2021physg} are the baseline methods of PANDORA~\cite{dave2022pandora}, InvRender~\cite{zhang2022invrender} accounts for indirect lighting in the BRDF estimation and employs the Spherical Gaussian to represent direct or indirect lighting. NVDiffRec~\cite{hasselgren2022shape} utilizes differentiable Monte-Carlo sampling sampling with a denoiser. Ref-NeuS~\cite{ge2023ref} aims to reduce ambiguity by attenuating the effect of reflective surfaces, while NeRO~\cite{liu2023nero} proposes to reconstruct the geometry and BRDF of objects with strong reflective appearances.

These methods typically rely on RGB data, which can struggle with accurate geometry reconstruction and radiance decomposition due to the limitations of using only intensity measurements. This often results in artifacts and inconsistencies, particularly in areas with strong specular reflections. We propose that incorporating polarization information is essential as it connects surface normals with lighting and material properties, improving the accuracy of these processes. Our evaluations, using open-source code from the original authors, indicate that our approach still delivers superior quality as shown in Tab.~\ref{tab:quality}. However, due to the inherent limitations in various methods, such as, PhySG's overly smooth geometry and inaccurate radiance decomposition, InvRender's superior performance only in synthetic scenarios, Ref-NeuS's effectiveness in view-dependent weighting scheme, and NeRO's proficiency in handling strong reflective objects, conventional objects in real-world settings often exhibit \emph{sub-optimal} performance. 

In Tab.~\ref{tab:quality}, we further conduct a thorough evaluation concerning the quantitative accuracy on the aforementioned test set. Firstly, we assess the rendering quality of our method and compare it to state-of-the-art algorithms. Hereby, we report the average PSNR and SSIM in comparison to the ground-truth test images. Next, we employ an invasive method to reconstruct the ground-truth shapes for these highly specular objects, so as to facilitate numerical assessment on the geometry recovery. Specifically, we apply a diffuse developer to objects and scan them using a high-end industrial-level 3D scanner. However, due to the potential inconsistency between the scanning and reconstruction coordinate systems, we manually scale and translate the scanned model to align with the reconstruction coordinate system. Subsequently, we utilize the non-rigid ICP algorithm to achieve the complete alignment between the scanned model and the reconstructed model under the shared coordinate system. Once aligned, the sum of the bi-directional chamfer distance (CD) between the reconstructed and scanned models is computed.

\begin{figure}[tp!]
    \centering
    \includegraphics[width=\columnwidth]{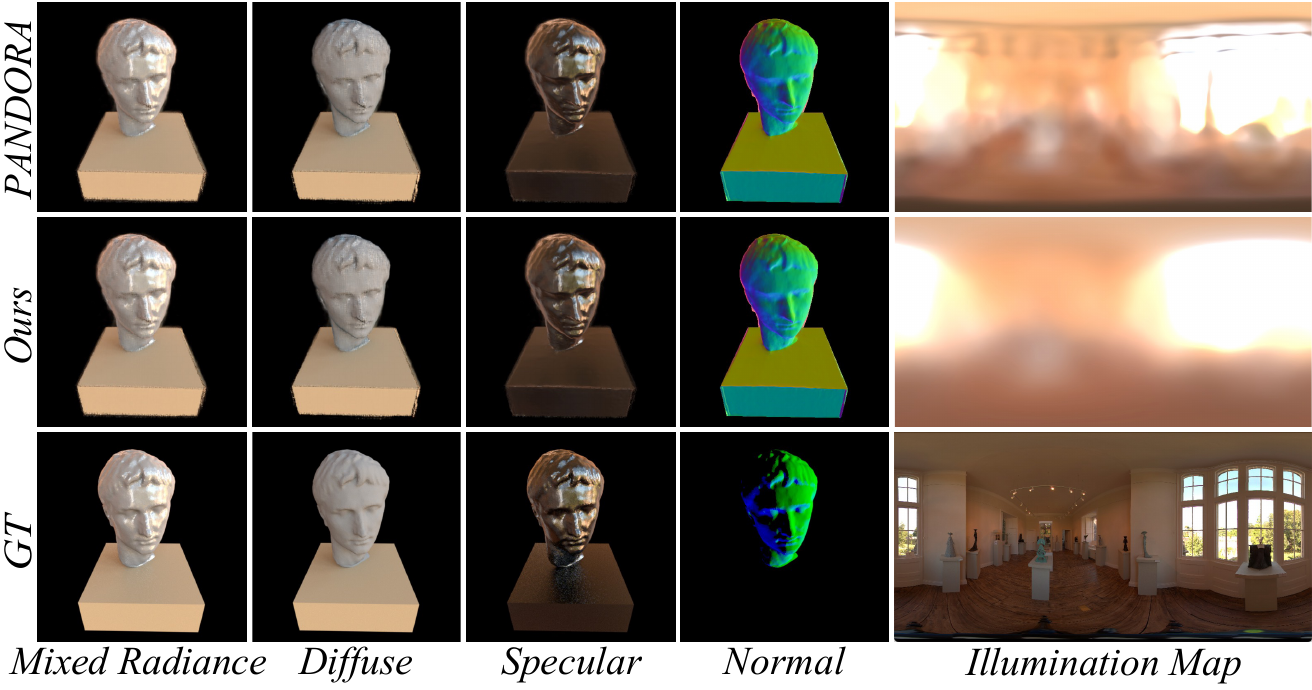}
    \vspace{-12pt}
    \caption{\textbf{Comparison of reflectance separation and surface normals with baselines on rendered \textit{Bust} model.} Note that, although PANDORA outputs sharp results, our method is also able to produce comparable results, because overall we use fewer constraints and need to solve for more unknowns.}
    \label{fig:results-syn}
    \vspace{-3pt}
\end{figure}

As depicted in Fig.~\ref{fig:comp-sota}, the outcomes indicate that NVDiffRec encounters challenges in effectively disentangling the diffuse and specular components and yields fundamentally erroneous geometric estimations. Surprisingly, this deficiency appears to exert minimal influence on the ultimate rendering quality, as evidenced by the high PSNR and SSIM metrics. We hypothesize that this arises from the method's inability to effectively resolve the inherent ambiguity between these two components, yet it still manages to yield exceptional rendering results grounded primarily in RGB loss. Conversely, NeRO exhibits improved geometric reconstruction capabilities, but its performance in radiance decomposition is lackluster. This arises from its rigid design tailored for entirely specular objects.

\begin{table}[t]
\caption{\textbf{Quantitative evaluation on rendered \textit{Bust} model.} We evaluate our method and PANDORA on $10\%$ held-out testsets of 45 images, and report the average peak signal-to-noise ratio (PSNR) and structured similarity (SSIM) of diffuse, specular and mixed radiance, mean angular error (MAE).}
\label{tab:quanti-syn}
\vspace{-6pt}
\centering
\resizebox{\columnwidth}{!}{
\begin{tabular}{c|cc|cc|cc|c}
\hline
\multirow{2}{*}{Method} & \multicolumn{2}{c|}{Diffuse} & \multicolumn{2}{c|}{Specular} & \multicolumn{2}{c|}{Mixed} & Normals \\ \cline{2-8} 
 & \multicolumn{1}{c|}{PSNR~$\uparrow$} & SSIM~$\uparrow$ & \multicolumn{1}{c|}{PSNR} & SSIM & \multicolumn{1}{c|}{PSNR} & SSIM & MAE~$\downarrow$ \\ \hline
PANDORA & \multicolumn{1}{c|}{\textbf{23.97}} & \textbf{0.907} & \multicolumn{1}{c|}{\textbf{26.02}} & \textbf{0.864} & \multicolumn{1}{c|}{\textbf{26.86}} & \textbf{0.895} & \textbf{4.096$^\circ$} \\
Ours & \multicolumn{1}{c|}{23.29} & 0.887 & \multicolumn{1}{c|}{25.97} & 0.860 & \multicolumn{1}{c|}{26.53} & 0.888 & 4.227$^\circ$ \\ \hline
\end{tabular}
}
\vspace{-6pt}
\end{table}

\paragraph{Comparisons with polarization methods.}
Tested on the synthetic data, both visualized results (Fig.~\ref{fig:results-syn}) and quantitative comparison (Tab.~\ref{tab:quanti-syn}) reveal that our method achieves comparable performance with SOTAs. Using \textit{Bust} model as an example, we present the ground-truth diffuse and specular components, as well as normals and environment map.

\begin{figure}[tp!]
    \centering
    \includegraphics[width=\columnwidth]{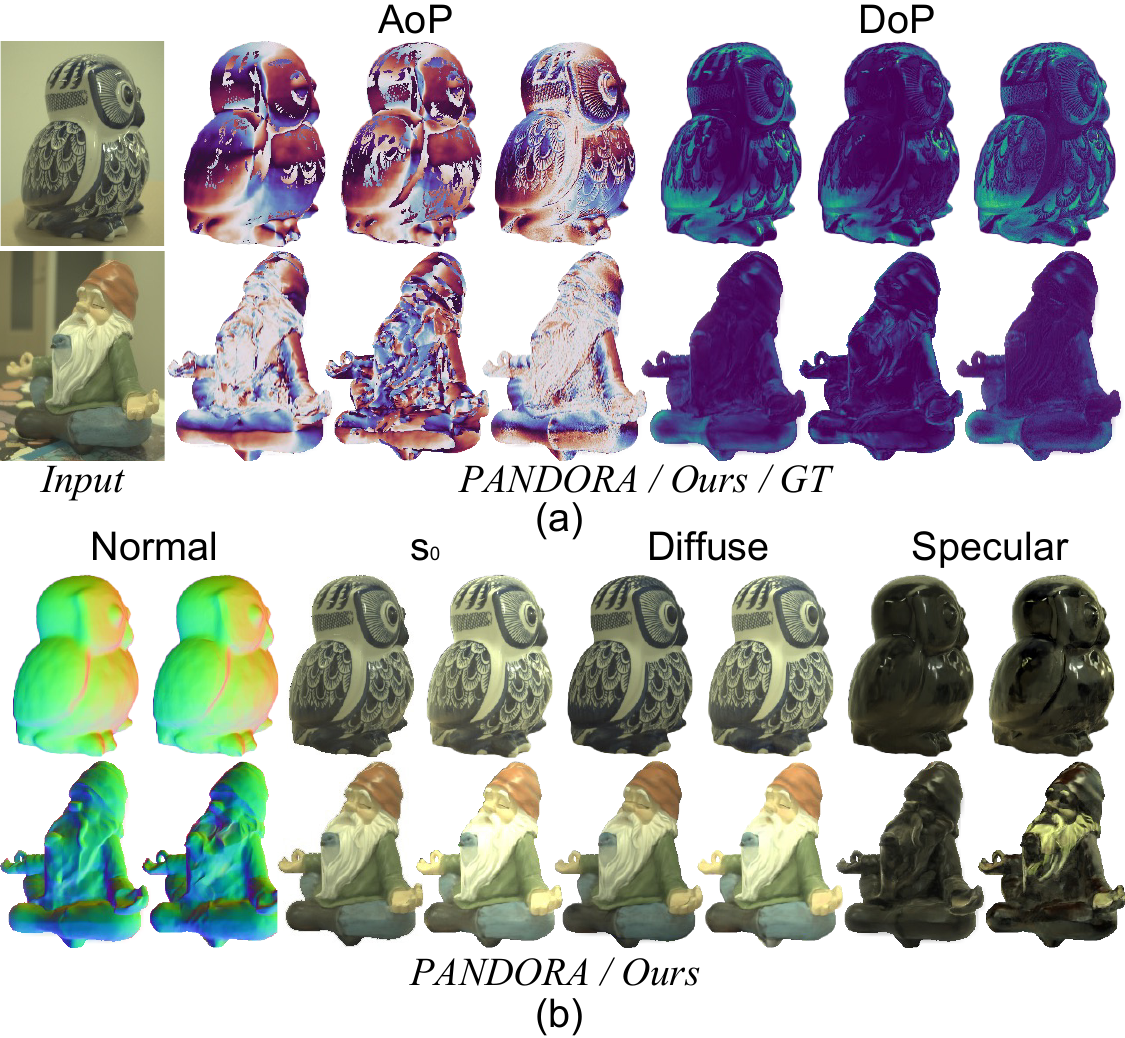}
    \vspace{-9pt}
    \caption{\textbf{Results of \textit{Owl} and \textit{Gnome} models.} (a) Comparison of the estimated AoP and DoP. (b) Comparison of the estimated geometry and radiance decomposition. For \textit{Owl} model, the average PSNR/SSIM on $10\%$ held-out test set between the estimated results of $s_0$ and the corresponding ground-truth are $24.46/0.8756$ (ours) and $25.07/0.8972$ (PANDORA). The PSNR/SSIM of which on \textit{Gnome} model are $28.13/0.9274$ and $28.43/0.9378$.}
    \vspace{-6pt}
    \label{fig:comp-pandora-all}
\end{figure}

Next, we study the raw data collected by PANDORA~\cite{dave2022pandora} (\textit{Owl} and \textit{Gnome}), as shown in the leftmost column of Fig.~\ref{fig:comp-pandora-all}(a). These datasets are obtained by acquiring raw images with a dedicated polarization camera equipped with SONY IMX250MZR sensor~\cite{sony}. After demosaicing, the raw image could be decomposed into four polarization images with different polarizing angles of $0^{\circ}$, $45^{\circ}$, $90^{\circ}$, and $135^{\circ}$.
In the following experiments, we use the image with the polarizer's angle of $135^{\circ}$ as input and leverage our approach to implicitly reconstruct the Stokes vectors and other information. For each case, we randomly select $90\%$ of the images for training. The results are shown in Fig.~\ref{fig:comp-pandora-all}.

It is noteworthy that, in PANDORA, the AoP and DoP are directly calculated from the captured data and are used as ground truths. In contrast, our approach generates intermediate outputs from the network, and our results can also nicely interpret the polarization states. Furthermore, since polarization is closely related to surface geometry and material properties, better estimated polarization cues result in high-quality decomposed diffuse and specular components, such as the tummy of the \textit{Owl} and the beard of the \textit{Gnome}. 

\subsection{Analysis}

\paragraph{Ablation study.}

\begin{figure}[t!]
    \centering
    \includegraphics[width=\columnwidth]{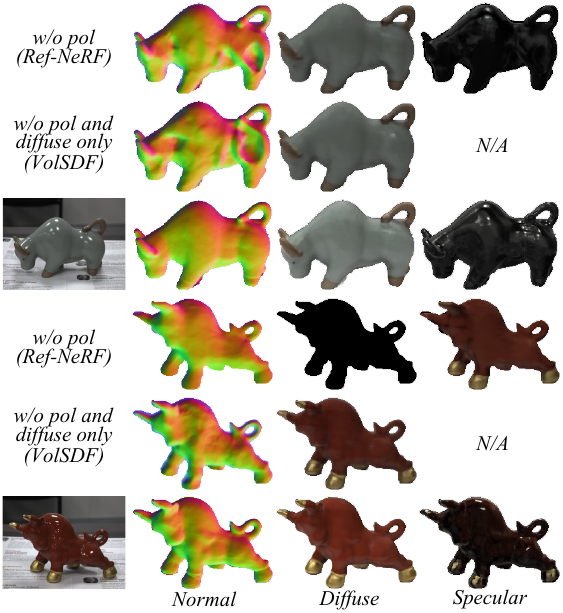}
    \vspace{-6pt}
    \caption{\textbf{Ablation study.} For each example, the top row depicts the results obtained by excluding polarization cues during rendering. Additionally, we exclusively focus on the diffuse components, and the corresponding outcomes are presented in the middle row. The bottom row showcases our outputs.}
    \label{fig:ablation}
        \vspace{-6pt}
\end{figure}

As shown in Fig.~\ref{fig:ablation}, we conduct two ablation studies for validation, such as, the effectiveness of polarization cues and the consideration of specular components. We first replace the polarized rendering as described in Sec.~\ref{subsec:pol_rendering} with the normal volume rendering. This design choice is actually an enhanced variant of Ref-NeRF~\cite{verbin2022ref}. Secondly, we compute the RGB loss between the rendered diffuse radiances, by removing the specular component during rendering, and the ground truth, this is actually VolSDF~\cite{yariv2021volume} with mask supervision used as a baseline.

\begin{figure*}[t]
    \centering
    \includegraphics[width=\textwidth]{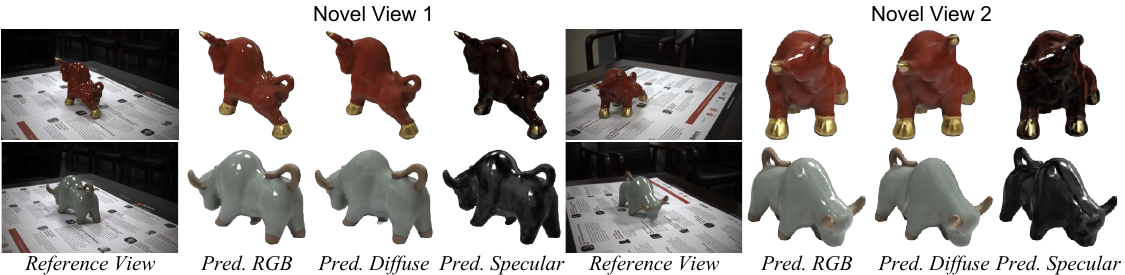}
    \vspace{-12pt}
    \caption{\textbf{Novel view synthesis results} of real-world captured objects. Remarkably, despite never encountering this particular perspective during training, the network is still capable of producing reasonably accurate rendering results.}
    \vspace{-6pt}
    \label{fig:nvs}
\end{figure*}

\begin{figure}[t!]
    \centering
    \includegraphics[width=0.94\columnwidth]{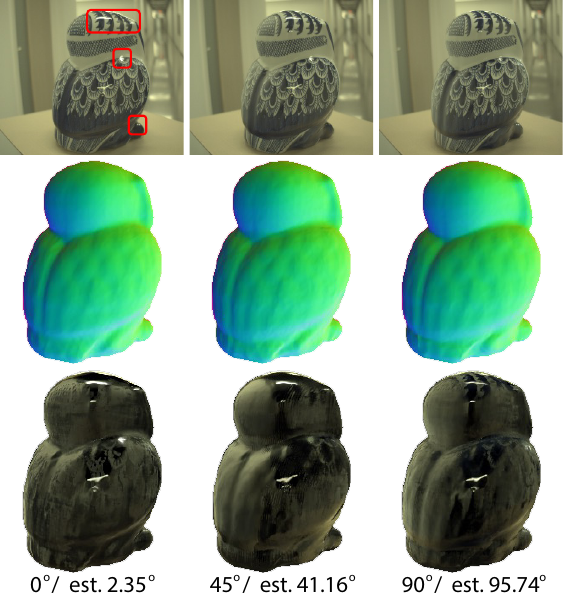}
    \caption{\textbf{Robustness analysis.} Despite minor color variations occur in specular regions across different polarization angles, particularly those highlights indicated by red boxes, our algorithm effectively restores a coherent geometry, while accurately recovers the corresponding specular map.}
    \label{fig:robustness}    \vspace{-6pt}
\end{figure}

Based on the reconstruction results, such as the top case of Fig.~\ref{fig:ablation}, where the surface exhibits distinct specular regions, without polarization cues, the network faces challenges in accurately learning distinctive features, leading to less precise surface geometry.
Despite this, our method still demonstrates robustness in capturing surface details, even in regions with prominent specular components. 

On the other hand, the final radiance decomposition results demonstrate that polarization cues can aid the network in better approximating true diffuse and specular components. In general, to ensure the consistency across multiple views, the network tends to focus on learning the diffuse components. As depicted, in the absence of polarization information, the network lacks substantial physical constraints, making it challenging to learn results that adhere to physics principles. In contrast, our method faithfully follows the polarization theorem during the rendering process, enabling more intuitive and reasonable decomposition.

\paragraph{Novel view synthesis.}
We conduct experiments on a held-out test set of the captured objects. These images are carefully chosen to be distinct from the existing viewing angles in the training set. During the testing phase, the network automatically generates essential information, including surface normals, polarization states, and decomposed radiances, using only the provided camera poses. The rendered visualizations of our results are illustrated in Fig.~\ref{fig:nvs}.

\paragraph{Robustness to different angles of the polarizer.}
As previously mentioned, our approach does not require the polarization angle of the input image to be calibrated in advance, as this information can be implicitly solved by the network. From another perspective, the network itself is ignorant of the polarization angle of the input image, and we can theoretically obtain the same reconstruction results. To verify this, we synthesize images with different polarization angles, such as $0^\circ$, $45^\circ$, and $90^\circ$, using Eq.~\ref{eq:pol}, as shown in Fig.~\ref{fig:robustness}. Our algorithm produces consistent and high-quality reconstruction results for different inputs. In addition, we output the estimated angle of the polarizer from the network, with an error less than $5^\circ$.

\section{Discussion and Conclusion}
This work presents advancements in polarization-based 3D reconstruction of glossy objects, by tackling the highly challenging yet novel task of estimating geometry and appearance from multi-view images with one single polarization angle per-view without pre-calibration. 
We introduce a fully differentiable polarization rendering pipeline that streamlines data acquisition to a single image per view and automatically determines the polarizer angle, eliminating manual calibration requirements and reducing costs. 

Despite challenges such as color bleeding, our approach accurately reconstructs object geometry and material properties, predicting diffuse and specular maps essential for polarization cues. 
By implicitly estimating the polarization angle to render a polarized image and comparing it to the captured image to compute loss, our integration of polarization information reinforces the relationship between surface normals and radiances, facilitating precise estimation of components for accurate geometry reconstructions. This work paves the way for high-fidelity reconstruction using accessible tools, with potential applications on devices like smartphones or IoTs.

\section*{Acknowledgments}
This work was partially supported by NSFC (U24B20154, 62322217, 62322207), Ant Group, Information Technology Center and State Key Lab of CAD\&CG, Zhejiang University, and  the Research Grants Council of Hong Kong (ECS 27212822, GRF 17208023).

{
    \small
    \bibliographystyle{ieeenat_fullname}
    \bibliography{main}
}

\clearpage
\maketitlesupplementary

\section{Neural Radiance Field}\label{sec:nrf}

\paragraph*{Neural implicit surface.}

We apply the neural volume rendering framework to represent implicit surfaces and follow VolSDF~\cite{yariv2021volume} to parameterize the density values with the transformation of an SDF.
For each pixel, we sample $N$ points along the camera ray and approximate the color $\hat{C}$ by:
\begin{equation}
    \hat{C} = \sum_{i=1}^N w_i c_i,
    \label{eq:volsdf_1}
\end{equation}
\begin{equation}
    \text{with}~w_i = T_i \left( 1 - \exp(-\sigma_i \delta_i) \right), ~T_i = \exp\left( -\sum_{j=1}^{i-1} \sigma_j \delta_j \right),
    \label{eq:volsdf_2}
\end{equation}
where $w_i$ is the weight of rendering, $\sigma_i$ and $c_i$ denote the density and color at each sampled point $i$ on the ray, and $\delta_i$ is the distance between adjacent samples. The density is defined as Laplace’s cumulative distribution function applied to a signed distance $d$, as follows:
\begin{equation}
\sigma(d)=\left\{\begin{matrix}
\frac{1}{2\beta}\exp(\frac{d}{\beta}) & if~d \leq 0 \\
\frac{1}{\beta}\left( 1 - \frac{1}{2}\exp(1-\frac{d}{\beta})) \right)      & if~d > 0
\end{matrix}\right..
\label{eq:lcd}
\end{equation}

Herein, $\beta$ is a learnable parameter during network training. In practice, we use MLPs to take 3D coordinates as input and output the corresponding signed distance as well as a global geometric feature vector. Referring to Eq.~\ref{eq:lcd}, the estimated SDFs are transformed to density values for volumetric integration of Eq.~\ref{eq:volsdf_2}.

\paragraph{Decomposed radiance fields.}
The outgoing radiance $c$ of a sampled point $\mathbf{x}$ on the camera ray can be decomposed into diffuse radiance $c^d$ and specular radiance $c^s$, respectively, as follows: \begin{equation}
    c^d = f_\theta(\mathbf{b}, \mathbf{x}), \;
    c^s = g_\theta(\mathbf{b}, \mathbf{IDE}(\eta, \omega_r)), \;
     \text{and} \; c = \gamma(c^d + c^s),
\label{eq:rad_decomp}
\end{equation}
where $f_\theta(\cdot)$ and $g_\theta(\cdot)$ denote MLPs with learnable parameters, and $\mathbf{b}$ is the geometric feature vector as mentioned above. Following the representations in Eq.~\ref{eq:pbrdf}, the diffuse surfaces should satisfy the property of Lambertian, thus $c^d$ in fact is only a function of position. However, for spatially-varying specular effects, following Verbin~\etal~\cite{verbin2022ref}, the radiance has strong correlations with surface roughness $\eta$ and the reflective direction of light $\omega_r$. With integrated directional encoding (IDE), the directions are encoded with a set of spherical harmonics, which enables the network to better reason about the inherent properties of the material. Finally, the diffuse and specular components are combined together with a fixed tone mapping function $\gamma$.

\section{Polarimetric BRDF Model}\label{sec:pbrdf_model}

In this work, we only consider linear polarization and build a scalable setup for the polarization image acquisition. To provide a clearer understanding of how polarization information is utilized in our method, We begin by presenting in the following the fundamental concepts.

\paragraph{Stokes vector.}
The polarization state of light is often characterized by the Stokes vector $\mathbf{s}$, which is usually computed by taking a series of measurements with different rotation angles, for example, polarized images with four different polarizing angles $0^{\circ}$, $45^{\circ}$, $90^{\circ}$ and $135^{\circ}$, represented by $\mathbf{I}_0$, $\mathbf{I}_{45}$, $\mathbf{I}_{90}$ and $\mathbf{I}_{135}$:
\begin{equation}
\mathbf{s} = [s_0,s_1,s_2,s_3]^T = [\mathbf{I}_0+\mathbf{I}_{90},\mathbf{I}_0-\mathbf{I}_{90},\mathbf{I}_{45}-\mathbf{I}_{135},0]^T.
\label{eq:comp_stokes}
\end{equation}

\paragraph{Mueller matrix.}
Any change of the polarization state due to the interaction with optical elements, such as linear polarizers or object surfaces, can be denoted as a multiplication of the corresponding Stokes vector with a Mueller matrix $\mathbf{M} \in \mathbb{R}^{4 \times 4} $. The incident and outgoing Stokes vector, represented by $\mathbf{s}^\text{in}$ and $\mathbf{s}^\text{out}$, respectively, are related by
\begin{equation}
    \label{eq:mueller}
    \mathbf{s}^\text{out} = \mathbf{M} \mathbf{s}^\text{in}.
\end{equation}

For surface reflection, considering the distant incident illumination $L_i$, which is commonly assumed to be unpolarized, its corresponding Stokes vector is denoted as $\mathbf{s}_i=L_i[1,0,0,0]^T$. Based on the pBRDF model proposed by  Baek~\etal~\cite{baek2018simultaneous}, the Mueller matrix can be decomposed as the sum of  diffuse component $\mathbf{M}^d$ and specular component $\mathbf{M}^s$, i.e., $\mathbf{M} = \mathbf{M}^d + \mathbf{M}^s$. Therefore, the outgoing Stokes vector can be reformulated as follows:
\begin{equation}
\resizebox{0.92\hsize}{!}{$%
\mathbf{s}^\text{out} = (\mathbf{M}^d + \mathbf{M}^s)\mathbf{s}^\text{in} = \underbrace{L_i k_d (\mathbf{n}\cdot\mathbf{i})}_{c^d} \begin{bmatrix}
T_o^+ T_i^+ \\ T_o^- T_i^+ \beta_o \\ -T_o^- T_i^+ \alpha_o \\ 0
\end{bmatrix}
+ \underbrace{L_i k_s \frac{DG}{4(\mathbf{n}\cdot\mathbf{o})}}_{c^s} \begin{bmatrix}
    R^+ \\ R^- \gamma_o \\ -R^- \chi_o \\ 0
\end{bmatrix}.
$}
\label{eq:pbrdf}
\end{equation}

In essence, $\mathbf{M}^d$ and $\mathbf{M}^s$ depend on surface albedo, surface normals, refractive index, and lighting conditions. In short, $\mathbf{n}$, $\mathbf{i}$, and $\mathbf{o}$ represent surface normal, incident and outgoing light direction, respectively. $k_d$ is the diffuse albedo, $k_s$ is the specular albedo, $T$ and $R$ are the Fresnel transmission and reflection coefficients. Refer to Baek~\etal~\cite{baek2018simultaneous} for detailed explanations and computation of remaining parameters. Herein, we denote the coefficients of the two terms on the right side of Eq.~\ref{eq:pbrdf} as diffuse radiance $c^d$ and specular radiance $c^s$.

\section{Polarization Rendering}\label{subsec:pol_rendering}
As shown in Fig.~2 and the following rendering pipeline, we use a polarization image $\mathbf{I}_{\phi_\text{pol}}$ as the input and leverage polarimetric BRDF model, characterized by the neural radiance field, to estimate the outgoing Stokes vectors $\textbf{s}^\text{out}$, which lay the foundation for polarization rendering. Refer to Eq.~\ref{eq:pbrdf} in the supp. for how to render $\textbf{s}^\text{out}$ using diffuse, specular, and roughness components. 
Subsequently, we present a differentiable processing pipeline to estimate the $\phi_\text{pol}$, eliminating the need for precise polarization angle measurements and facilitating the implicit rendering of desired polarized images $\mathbf{I}^\text{out}_{\phi_\text{pol}}$ for loss calculation.
\begin{figure}[h]
    \includegraphics[width=\linewidth]{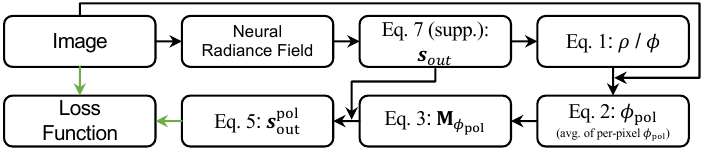}
\end{figure}

\section{Implementation Details}\label{subsec:impl}
The SDF network takes the 3D coordinate as input and applies the positional encoding (PE) to spatial locations using 6 frequencies. This encoded input is then processed through 8 fully connected layers with 256 channels each, utilizing ReLU activations. Additionally, the encoded input vector is connected to the output feature at the 4$^{th}$ layer through a skip connection. The network outputs the signed distance value and an extra 256-dimensional geometric feature vector. Notably, surface normals can be obtained as the normalized gradient of the neural SDF. To initialize parameters of the SDF network, we utilize geometric initialization methods as described by Gropp~\etal~\cite{gropp2020implicit}.

The diffuse radiance $f_{\theta}$, roughness, and mask prediction functions share similar network architectures. They take the concatenation of the geometric feature vector and the encoded spatial locations with 10 frequencies as input. The network is composed of 4 MLP layers with a width of 512 channels. The output structures contain 3 channels with \textit{sigmoid}, 1 channel with \textit{softplus}, and 1 channel with \textit{sigmoid}, respectively. For the estimation of specular components~\cite{verbin2022ref}, we enable the network to reason about radiances with the integrated directional encoding of roughness and the encoded reflective directions with PE of 2 frequencies. $g_{\theta}$ also uses 4 fully connected MLP layers with 512 channels per layer and outputs 3 channels with the \textit{softplus}.

Our algorithms are implemented in Pytorch~\cite{paszke2019pytorch}. In our experiments, we use a batch size of 512 rays, each sampled at 128 locations. We use the Adam optimizer~\cite{kingma2014adam} ($\beta_1=0.9$, $\beta_2=0.999$) with a learning rate that begins at $5 \times 10^{-4}$ and decays exponentially to $5 \times 10^{-5}$ during training. To better warm up the training, in the early 10k iterations, we define $\mathcal{L}_\text{rgb}$ as the loss between the predicted radiance $c$ in Eq.~\ref{eq:rad_decomp} and the ground truth. In the next 5k iterations, we replace $c$ with the diffuse components of $\mathbf{s}^\text{out}_{\phi_\text{pol}}$, which are subsequently used for loss computation. In addition, The refractive index of the object is set to 1.5. The optimization for a single object typically takes around 200k iterations to converge on a single NVIDIA Titan X GPU ($\sim$ 2 days).

\end{document}